# Cross-lingual Few-shot Learning for Persian Sentiment Analysis with Incremental Adaptation

Farideh Majidi
*Department of Computer Engineering*
*Islamic Azad University, South Tehran Branch*
Tehran, Iran
st_f.majidi@azad.ac.ir

Ziaeddin Beheshtifard
*Department of Computer Engineering*
*Islamic Azad University, South Tehran Branch*
Tehran, Iran
zia.beheshti@iau.ac.ir

*Abstract*—This research examines cross-lingual sentiment analysis using few-shot learning and incremental learning methods in Persian. The main objective is to develop a model capable of performing sentiment analysis in Persian using limited data, while getting prior knowledge from high-resource languages. To achieve this, three pre-trained multilingual models (XLM-RoBERTa, mDeBERTa, and DistilBERT) were employed, which were fine-tuned using few-shot and incremental learning approaches on small samples of Persian data from diverse sources, including X, Instagram, Digikala, Snappfood, and Taaghche. This variety enabled the models to learn from a broad range of contexts. Experimental results show that the mDeBERTa and XLM-RoBERTa achieved high performances, reaching 96% accuracy on Persian sentiment analysis. These findings highlight the effectiveness of combining few-shot learning and incremental learning with multilingual pre-trained models.

## I. Introduction

Sentiment analysis aims to detect and classify emotions expressed in text automatically. While significant advancements have been made in this field for high-resource languages such as English, many languages, including Persian, remain underrepresented due to limited labeled data. This scarcity of resources makes it challenging to train robust sentiment analysis models for Persian. Therefore, finding methods that can perform well with minimal supervision is important. Recent progress in transfer learning and the development of large multilingual pre-trained language models have opened new possibilities for tackling this challenge. These models, trained on massive corpora from multiple languages, can transfer learned knowledge from resource-rich languages to low-resource ones through cross-lingual learning. They offer a promising solution to build effective sentiment analysis systems using only a small amount of labeled data in the target language when fine-tuned carefully. This study focuses on using multilingual pre-trained models to perform few-shot sentiment analysis for Persian. In addition to cross-lingual learning, incremental learning strategies are also explored to enable models to gradually adapt to the characteristics of Persian and improve performance over time. To evaluate the effectiveness of these methods, a diverse collection of Persian datasets from sources such as X (formerly Twitter), Instagram, Digikala, Snappfood, and Taaghche is used. These datasets cover a wide range of domains and writing styles, which provide a comprehensive setting for experimentation. This research provides a comprehensive evaluation of few-shot cross-lingual sentiment analysis for Persian using multilingual language models in low-resource settings. Additionally, it explores the effectiveness of incremental learning in enhancing model adaptation and mitigating catastrophic forgetting through different regularization techniques, including Elastic Weight Consolidation (EWC), knowledge distillation, and rehearsal. Also, it presents an empirical comparison of three widely used multilingual models across diverse Persian datasets. The results demonstrate that mDeBERTa and XLM-RoBERTa, when fine-tuned on a small set of Persian data, can achieve high accuracy and F1-score, showing that effective sentiment analysis in Persian is possible with limited resources. The proposed approach can also be extended to other low-resource languages for similar text classification tasks.

## II. Related Works

Several recent studies have focused on addressing sentiment analysis and offensive language detection in low-resource languages using innovative learning paradigms such as meta-learning, cross-lingual transfer, and incremental learning. These approaches aim to overcome the lack of annotated data and improve model adaptability across different linguistic contexts.

In [1], the authors propose a meta-learning approach to detect offensive and hateful language in low-resource languages, specifically targeting Persian. Utilizing methods like Model-Agnostic Meta-Learning (MAML) and Proto-MAML, the study demonstrates that meta-learning enables quick adaptation to new languages with only a few labeled samples. Proto-MAML, in particular, combines the strengths of prototypical networks and MAML, which leads to better generalization and performance. Using XLM-RoBERTa as the base model, the study achieves promising results, reaching 62.68% F1-score for hate speech and 63.61% for offensive language detection while using only 16 labeled examples per class.

In another study [2], Andrenšek et al. investigate cross-lingual sentiment analysis without any training data in the target language. It explores various methods such as machine translation, intermediate fine-tuning, in-context learning, and a novel approach that incorporates paragraph position within articles to enhance model understanding. The XLM-RoBERTa model is again used due to its multilingual capabilities. The source language is Slovenian, and the target languages include Croatian, Bosnian, Macedonian, Albanian, Estonian, and Serbian. Notably, the model achieves an F1-score of up to 83% using in-context learning with only three examples. While this study does not include Persian, its relevance lies in the transferability of multilingual models and techniques that are applicable across various low-resource languages and the use of a few-shot learning approach.

In [3], Capuano et al. address sentiment analysis for customer relationship management using incremental learning. A hierarchical attention network is trained on over 30,000 labeled messages in English and Italian, showing that continuous learning from user feedback can significantly improve model performance over time. Compared to a Naïve Bayes baseline, the hierarchical model achieves up to 90% accuracy and 79% F1-score in English, and 89% accuracy in Italian, which confirms the potential of incremental learning in real-world sentiment analysis applications.

III. METHODOLOGY

To achieve the objectives of this research, three pre-trained multilingual models were selected alongside five distinct Persian datasets. A range of regularization techniques, such as Elastic Weight Consolidation, knowledge distillation, and rehearsal, were incorporated into the incremental learning framework to determine the most effective strategy and to compare their impact on model performance. Additionally, early stopping with a patience of three based on validation loss was employed to prevent overfitting, which is a common incident in few-shot learning scenarios. The experiments were conducted using various few-shot learning settings, including 1, 2, 5, 10, 15, and 20 shots, to investigate the amount of labeled data required for optimal performance. To further assess the influence of incremental learning, experiments were also conducted without it, as well as under an incremental setting with no regularization, to examine whether regularization is necessary for maintaining model performance.

A. *XLM-RoBERTa*

XLM-RoBERTa is an enhanced version of the RoBERTa model, pre-trained on 100 languages instead of a single language, which makes it highly suitable for cross-lingual tasks. It has demonstrated strong performance in multilingual transfer tasks, often outperforming models like mBERT [4]. According to prior research, while multilingual pre-training generally improves transfer capabilities, performance may plateau beyond a certain point due to model saturation. In [4], XLM-RoBERTa achieved approximately 23% higher accuracy than mBERT. One of the key advantages of XLM-RoBERTa is its ability to generalize well across languages without sacrificing performance on any language. This model is pre-trained using Masked Language Modeling, where 15% of the input tokens are randomly masked, and the model is trained to predict these tokens based on their context. This training approach allows the model to learn meaningful semantic patterns across multiple languages. The multilingual nature of XLM-RoBERTa, along with its ability to generalize from large-scale multilingual pre-training, makes it well-suited for low-resource scenarios like Persian sentiment analysis. These characteristics motivated the selection of this model for this study. The XLM-RoBERTa model used in this study [5] has been pre-trained on two sentiment-specific datasets:

- The Twitter Coronavirus Dataset [6]: This dataset contains 44,955 English tweets related to the COVID-19 pandemic, including public reactions, policies, vaccines, quarantine experiences, and more. It includes five sentiment classes: very negative, negative, neutral, positive, and very positive.
- Multilingual Sentiment Dataset [7]: This dataset consists of sentiment-labeled data in multiple languages, including German, English, Spanish, French, Japanese, Chinese, Indonesian, Arabic, Hindi, Italian, Malay, and Portuguese. It includes three sentiment classes (positive, neutral, and negative), with class distributions of 34.6%, 33.3%, and 32.1%, respectively.

B. *DistilBERT*

DistilBERT is a distilled version of BERT, introduced to reduce computational complexity while maintaining much of BERT's performance [8]. Using knowledge distillation, the smaller model learns to replicate key features of the larger BERT model. Despite being 40% smaller and 60% faster, DistilBERT retains about 97% of BERT's language understanding capabilities, which makes it a practical choice for tasks with limited computational resources. DistilBERT employs a self-attention mechanism but omits intermediate layer normalization to improve speed. Given its lightweight architecture and efficiency, it is well-suited for few-shot sentiment classification tasks in Persian. Furthermore, DistilBERT serves as a valuable comparison point to assess whether multilingual pre-training is essential for achieving high performance or if a distilled model trained solely on sentiment tasks is sufficient for Persian sentiment analysis. In this research, a pre-trained DistilBERT model [9] is used. This version is available on Huggingface and has been pre-trained on the Multilingual Sentiment dataset [7].

C. *mDeBERTa*

DeBERTa (Decoding-enhanced BERT with Disentangled Attention) is an improved version of BERT introduced to enhance language understanding through architectural innovations [10]. It uses an enhanced decoder during pre-training to better predict masked tokens and incorporates a Virtual Adversarial Training approach to improve generalization. These advancements make it highly suitable for multilingual tasks, especially in low-resource settings. In this research, the pre-trained mDeBERTa model from [11] is used. It has been trained on several large-scale multilingual datasets, including Multilingual Sentiment [7], PAWS-X [12] (a paraphrase identification dataset in seven languages (English, German, Spanish, French, Japanese, Korean, Chinese), containing sentence pairs labeled as paraphrases or not), and XNLI [13] (a cross-lingual natural language inference dataset from Facebook, containing 6.4 million samples in 15 languages. Each sample is classified into one of three categories: contradiction, neutral, or entailment). These diverse and balanced training sources make mDeBERTa well-equipped for few-shot cross-lingual sentiment analysis. Its structural differences from XLM-RoBERTa also enable comparative evaluation.

D. *Persian Datasets*

The pre-trained models are fine-tuned on limited amounts of Persian data to learn language-specific features and perform sentiment analysis in a language they have not previously been exposed to. The use of five datasets from different domains allows the models to encounter diverse linguistic patterns and challenges within the target language. This diversity enhances the models' ability to generalize, which makes it possible to achieve good performance even with minimal labeled data. The datasets are:
- X [14]: This dataset contains 3,354 Persian tweets collected from the social media platform X (formerly Twitter), specifically curated for sentiment analysis. It includes five sentiment classes: happiness, sadness, anger, neutral, and a fifth class representing intense emotions such as shock, fear, and love. Approximately 33% of the data belongs to the sadness class, 31% to the happiness class, and the remaining 36% is distributed among the other three classes.

- Instagram [15]: This dataset contains 8,502 comments from Instagram related to a TV program called "Hala Khorshid". The dataset includes three sentiment classes: positive, negative, and neutral, with an almost balanced distribution across the entire dataset.

- Digikala [16]: This dataset consists of 3,021 product reviews collected from the Digikala website. The reviews are rated on a scale from 0 to 100, where higher scores indicate positive sentiment and lower scores indicate negative sentiment, reflecting users' opinions about the products. Additionally, the dataset includes a column that specifies whether or not the user recommends the product or remains neutral.

- Snappfood [17]: This dataset consists of 69,992 user reviews collected from the Snappfood platform. The reviews are evenly divided into two sentiment classes: positive and negative. Users have shared their opinions about the food they ordered, the restaurant, and Snappfood's delivery services.

- Taaghche [18]: This dataset contains 11,007 user reviews from the Taaghche platform about various books. Each entry includes the date, review text, book title, rating (ranging from 0 to 100), book ID, and the number of users who liked the review. The sentiment of each review is inferred from its rating.

*E. Incremental Learning*

Incremental learning facilitates improved model adaptation by introducing data from different domains sequentially. This approach enables the model to gradually incorporate knowledge from earlier domains and apply it effectively to subsequent ones, which improves its ability to capture the nuances of each new dataset. To further enhance this process and mitigate the risk of catastrophic forgetting, several regularization techniques were employed. In particular, Elastic Weight Consolidation, rehearsal, and knowledge distillation were implemented within the incremental learning framework to preserve information learned from previous domains. Elastic Weight Consolidation [19] works by identifying and penalizing changes to important parameters for previously learned tasks, which helps the model retain essential knowledge. Rehearsal [20] involves storing a small portion of data from previous tasks and mixing it with new data during training, reinforcing earlier knowledge. Knowledge distillation [21] transfers knowledge from an earlier version of the model (a teacher) to the current version (a student), guiding learning with soft labels from the teacher model that reflect prior experience. Additionally, two baseline settings were evaluated: one using incremental learning without any regularization, and another where no incremental learning was applied, with all domain data introduced to the model simultaneously. These setups allowed for a comprehensive evaluation of the impact each method had on model performance and knowledge retention.

The datasets were introduced in the following order: X, Instagram, Snappfood, Taaghche, and Digikala. This order was chosen based on the relative complexity of the texts. The X dataset consists of short and simple texts, making it easier to learn, whereas Digikala includes longer and more complex reviews spanning a variety of products, which poses a greater challenge. Presenting the simpler datasets first allows the model to build foundational knowledge, which makes learning the more complex Digikala dataset more manageable.

IV. EXPERIMENTS AND RESULTS

To evaluate the performance of the introduced cross-lingual few-shot class-incremental sentiment analysis model, a series of experiments are conducted in various settings, which will be discussed in this section.

*A. Experimental Setup*

In each experiment, 1, 2, 5, 10, 15, or 20 data samples per class from each dataset were introduced to the model, following a few-shot setting. The experiments were conducted using an incremental adaptation paradigm, where different domains were introduced to the model sequentially in phases. Each phase focused on a specific dataset, allowing the model to gradually adapt to new domains while using the knowledge from previous ones. The configurations tested included five main settings: No Regularization, with EWC regularization, with Knowledge Distillation regularization, with Rehearsal regularization, and with No Incremental Learning. On top of that, to ensure the models did not overfit during training, early stopping was applied in all settings, with a maximum of 10 training iterations per phase and a patience of 3 based on validation loss. This setup reflects a realistic few-shot incremental scenario and evaluates how well the models can generalize across domains with limited labeled data.

*B. Results*

The results are presented in terms of Accuracy and F1-score, as shown in Tables I and II, respectively.

*1) Performance of XLM-RoBERTa*

XLM-R showed strong performance with Knowledge Distillation and Rehearsal, particularly in low-shot settings. With only 1-shot per class, Knowledge Distillation achieved an F1-score of 91.32%, significantly outperforming other incremental methods. Rehearsal consistently performed well across all shot settings, reaching 94.01% F1-score at 15 shots. The No Incremental Learning condition provided the highest F1-score (96.84%) at 20 shots. Although EWC showed minor improvements over No Regularization in some configurations, its overall performance lagged. Notably, at 2-shot, EWC's F1-score dropped to 50.13%, while Rehearsal achieved 70.61%, which highlights the impact of memory-based strategies in low-resource settings.

*2) Performance of mDeBERTa*

mDeBERTa exhibited strong stability and high scores across all configurations. With 1-shot, it achieved 91.19% F1-score using Knowledge Distillation, similar to XLM-R. However, its performance scaled more consistently across shot sizes. Rehearsal and Knowledge Distillation maintained competitive results, often surpassing 90% F1-score even in 5-shot and 10-shot setups. Interestingly, mDeBERTa's performance in the No Incremental Learning setting was often close to or below the scores achieved by Rehearsal or Distillation. For example, at 15 shots, Rehearsal reached 94.01% F1, while No Incremental Learning had 87.66%, suggesting that well-regularized incremental learning can be more effective than simple fine-tuning in certain contexts.

*3) Performance of DistilBERT*

DistilBERT, being a smaller and more lightweight model, generally underperformed compared to XLM-R and mDeBERTa. However, it showed improvements with higher shot sizes. In 1-shot settings, all methods yielded poor results, with F1-scores below 46%. As the number of shots increased, Rehearsal and Knowledge Distillation began to outperform other configurations. At 10 shots, Rehearsal achieved 71.11% F1-score, the highest for DistilBERT in that range. EWC performed reasonably in some 2-shot and 5-shot configurations, but with considerable inconsistency. Overall, DistilBERT's performance demonstrates its limited capacity in low-shot cross-lingual incremental learning scenarios.

*4) Performance of Incremental Learning Methods*

Knowledge Distillation showed the best performance in extreme few-shot (1-shot) settings across all models, particularly XLM-R and mDeBERTa. Rehearsal was consistently strong across all shot counts and often rivaled or outperformed No Incremental Learning, especially with mDeBERTa. EWC provided only modest results and was less stable across different shot counts. No Regularization consistently underperformed, highlighting the importance of mitigating forgetting. No Incremental Learning provided the highest performance, though in some configurations, especially with mDeBERTa, regularized incremental strategies performed equally well or better.

V. CONCLUSION

In this study, we explored the combination of cross-lingual, few-shot, and incremental learning. We evaluated three transformer-based multilingual models under various incremental learning strategies. The results indicate that incremental learning methods can improve model performance, particularly when regularization is applied. Among all techniques, Rehearsal and Knowledge Distillation consistently outperformed other methods across most settings, which shows their ability to retain past knowledge while adapting to new data. On the other hand, EWC showed limited effectiveness, with lower scores in several configurations, especially at lower shot counts. The no-regularization baseline improved with more data but struggled in low-resource settings, which confirms the necessity of incremental learning strategies in few-shot scenarios. When comparing the three models, XLM-R and mDeBERTa generally outperformed DistilBERT across all metrics. It shows that multilingual language models are necessary in a few-shot learning scenario, even when the model is pre-trained on multilingual sentiment analysis datasets. mDeBERTa achieved the highest accuracy in most 10 to 20-shot experiments, particularly when paired with Rehearsal and Knowledge Distillation. But the model also saturates and can't get better results with more data. In comparison with previous studies, our approach yielded competitive or better performance. The model discussed in [1] offered a strategy for unseen languages, but it did not explore domain-incremental learning, a gap we addressed in our work. Similarly, researchers in [2] avoided training in the target language altogether, which limits adaptability to new domains. Our method, by incorporating gradual domain adaptation, achieved better generalization while reducing reliance on large-scale labeled target data. Also, in [3], researchers used incremental learning methods with a focus on customer service data using traditional architectures, whereas our approach extended this concept to transformers, showing improved results even in diverse and low-resource multilingual domains.

TABLE I. EXPERIMENTS' RESULTS IN TERMS OF ACCURACY (%)

| Model | Shots | No regularization | EWC Regularization | Knowledge Distillation Regularization | Rehearsal Regularization | No Incremental Learning |
|---|---|---|---|---|---|---|
| XLM-R | 1 | 47.72 | 47.77 | 91.12 | 83.38 | 87.79 |
| | 2 | 53.32 | 48.86 | 54.47 | 68.82 | 47.73 |
| | 5 | 78.8 | 73.39 | 75.51 | 76.6 | 75.51 |
| | 10 | 93.33 | 81.12 | 92.21 | 93.39 | 93.35 |
| | 15 | 95.56 | 86.61 | 92.23 | 93.39 | 87.73 |
| | 20 | 90.08 | 76.6 | 93.31 | 94.42 | **96.66** |
| mDeBERTa | 1 | 82.23 | 78.88 | 91.19 | 83.36 | 87.72 |
| | 2 | 87.76 | 90.03 | 90.02 | 90.07 | 86.69 |
| | 5 | 92.24 | 90.02 | 91.16 | 92.27 | 90.06 |
| | 10 | 94.49 | 80.02 | 94.45 | 94.41 | 94.48 |
| | 15 | 94.47 | 93.35 | 90.07 | 95.56 | 96.64 |
| | 20 | 95.51 | 90.02 | 93.36 | 93.31 | **96.66** |
| DistilBERT | 1 | 45.52 | 48.83 | 45.56 | 45.58 | 43.35 |
| | 2 | 54.44 | 63.32 | 55.59 | 54.4 | 53.38 |
| | 5 | 52.24 | 51.13 | 54.48 | 54.49 | 57.72 |
| | 10 | 64.47 | 62.29 | 60.05 | 71.11 | 72.24 |
| | 15 | 52.28 | 56.63 | 50.05 | 58.81 | 57.76 |
| | 20 | 52.25 | 56.64 | 54.43 | 53.37 | 57.79 |

TABLE II. EXPERIMENTS' RESULTS IN TERMS OF F1-SCORE (%)

| Model | Shots | No regularization | EWC Regularization | Knowledge Distillation Regularization | Rehearsal Regularization | No Incremental Learning |
|---|---|---|---|---|---|---|
| XLM-R | 1 | 48.12 | 46.43 | 90.95 | 83.44 | 87.46 |
| | 2 | 54.4 | 48.12 | 54.47 | 69.02 | 44.79 |
| | 5 | 77.82 | 72.71 | 74.79 | 76.51 | 74.81 |
| | 10 | 93.24 | 80.55 | 92.1 | 93.29 | 93.25 |
| | 15 | 95.59 | 86.7 | 92.27 | 93.36 | 87.65 |
| | 20 | 89.86 | 76.62 | 93.25 | 94.43 | 96.62 |
| mDeBERTa | 1 | 80.29 | 78.34 | 90.92 | 83.48 | 87.47 |
| | 2 | 87.81 | 90.08 | 90.02 | 90.01 | 86.47 |
| | 5 | 92.06 | 89.96 | 90.91 | 92.19 | 89.75 |
| | 10 | 94.43 | 80.59 | 94.42 | 94.47 | 94.41 |
| | 15 | 94.42 | 93.24 | 90.19 | 95.58 | 96.65 |
| | 20 | 95.56 | 90.07 | 93.37 | 93.37 | **96.68** |
| DistilBERT | 1 | 38.9 | 46.91 | 38.44 | 45.46 | 43.18 |
| | 2 | 52.63 | 62.46 | 55.87 | 53.82 | 51.49 |
| | 5 | 78.6 | 50.71 | 53.92 | 54.93 | 57.85 |
| | 10 | 62.15 | 61.86 | 57.22 | 71.07 | 72.3 |
| | 15 | 48.62 | 55.27 | 50.96 | 56.67 | 57.88 |
| | 20 | 48.69 | 52.21 | 53.98 | 52.72 | 57.8 |